%% file: acl_latex.tex
\pdfoutput=1

\documentclass[11pt]{article}

\usepackage[final]{acl}

\usepackage{times}
\usepackage{latexsym}

\usepackage[T1]{fontenc}

\usepackage[utf8]{inputenc}

\usepackage{microtype}

\usepackage[utf8]{inputenc}
\usepackage{graphicx}
\usepackage{amssymb}
\usepackage{amsmath}
\usepackage{esvect}
\usepackage{svg}
\usepackage{comment}
\usepackage{lipsum}

\usepackage{algpseudocode}
\usepackage{varwidth}

\usepackage{subcaption}
\usepackage{rotating}
\usepackage{graphicx}

\usepackage{algorithm}
\usepackage{tabularx}
\usepackage{algpseudocode}
\usepackage{algpascal}

\usepackage{booktabs, multirow} 
\usepackage{soul}
\usepackage{makecell}
\usepackage{threeparttable}
\usepackage{footnote}
\makesavenoteenv{tabular}
\usepackage[labelformat=simple]{subcaption}

\usepackage[pdf]{graphviz}

\usepackage[normalem]{ulem}
\useunder{\uline}{\ul}{}

\usepackage{xspace}

\newcommand\blfootnote[1]{%
  \begingroup
  \renewcommand\thefootnote{}\footnote{#1}%
  \addtocounter{footnote}{-1}%
  \endgroup
}

\newcommand{\dialograph}{\textsc{DialoGraph}\xspace}
\newcommand{\ours}{\textsc{Theta}\xspace}
\newcommand{\ourht}{\textsc{ht}\xspace}
\newcommand{\da}{\textsc{da}\xspace}

\newcommand{\cb}{\textsc{CraigslistBargain}\xspace}
\newcommand{\abcd}{\textsc{Abcd}\xspace}
\newcommand{\callcenter}{\textsc{Call Center}\xspace}
\newcommand{\cls}{\texttt{[CLS]}\xspace}
\newcommand{\septoken}{\texttt{[SEP]}\xspace}
\newcommand{\ptytoken}{\texttt{[PTY]}\xspace}
\newcommand{\usrtoken}{\texttt{[USR]}\xspace}
\newcommand{\agttoken}{\texttt{[AGT]}\xspace}

\input{Definitions.tex}

\title{Unsupervised Learning of Hierarchical Conversation Structure}

\definecolor{green}{RGB}{0,128,0}

\providecommand{\roy}[1]{
    {\protect\color{orange}{[Roy: #1]}}
}

\providecommand{\nascomment}[1]{
    {\protect\color{blue}{[Noah: #1]}}
}

\newcommand\ai{$^{\diamondsuit}$}
\newcommand\uw{$^\spadesuit$}
\newcommand\msr{$^{\heartsuit}$}
\newcommand\aspace{\hspace{.75em}}
  
 \author{
     Bo-Ru Lu\uw\aspace
     Yushi Hu\uw\aspace
     Hao Cheng\msr\aspace
     Noah A. Smith\uw\ai\aspace
     Mari Ostendorf\uw\aspace\\
     \uw University of Washington \aspace
     \msr Microsoft Research \aspace
     \ai Allen Institute for AI\\
     {\tt \{roylu,yushihu,ostendor\}@washington.edu}\\
     {\tt chehao@microsoft.com nasmith@cs.washington.edu}
}

\begin{document}
\maketitle
\begin{abstract}
Human conversations can evolve in many different ways, creating challenges for automatic understanding and summarization. Goal-oriented conversations often have meaningful sub-dialogue structure, but it can be highly domain-dependent. This work introduces an unsupervised approach to learning hierarchical conversation structure, including turn and sub-dialogue segment labels, corresponding roughly to dialogue acts and sub-tasks, respectively.
The decoded structure is shown to be useful in enhancing neural models of language for three conversation-level understanding tasks. Further, the learned finite-state sub-dialogue network is made interpretable through automatic summarization. \blfootnote{We release our code for experiments at \url{https://github.com/boru-roylu/THETA}.}
\end{abstract}

\input{intro3}
\input{method2}
\input{experiment}

\input{analysis}

\input{related2}
\input{conclusion}
\input{acknowledgement}
\input{limitation}

\input{ethical}

\bibliography{anthology,custom}
\bibliographystyle{acl_natbib}

\pagebreak
\appendix
\input{appendix}

\end{document}

%% file: intro3.tex
\section{Introduction}
\label{sec:intro}

Increasingly, language understanding applications involve conversational speech and text. Much attention has recently been directed at human-agent dialogue systems, including virtual assistants, interactive problem solving, and information seeking tasks (e.g., conversational question answering). However, automatic understanding of human-human conversations is also of interest for problems such as call-center analytics, conversation outcome prediction, meeting summarization, and human-agent interaction involving multiple people. The focus of this paper is on human-human conversation understanding.

Like written documents, goal-oriented conversations tend to have structure (openings, context setting, problem solving, etc.). However, in human-human conversations (both text and speech), participant roles factor into the structure, and the structure is less rigid due to the need to accommodate miscommunications and varying objectives.
Yet, most work on conversational systems treats dialogues like written text, i.e., the dialogue history is a linear sequence of text. In this paper, we explore unsupervised learning strategies for adding structural information to a state-of-the-art hierarchical transformer-based model of text.

Linguistic analysis of conversations often involves associating speaker utterances with dialogue acts (DAs), e.g., question, statement, backchannel, clarification, etc. \cite{Jurafsky1997SWDA,Core97codingdialogs}, and segmenting the conversation into nested subsequences of participant turns that reflect a common topic or conversational goal \cite{grosz-sidner-1986-attention}. Past studies have explored using such structure, particularly DAs, to improve automated human-agent dialogues. Here, we use hierarchical structure (both turn-level DA labels and sub-dialogue states) to improve classification of human-human conversations. Specifically, we introduce \textbf{T}hree-stream \textbf{H}i\textbf{e}rarchical \textbf{T}r\textbf{a}nsformer (\ours), which integrates
transformer representations of the DA and sub-dialogue state sequences into a hierarchical transformer 
(\ourht) \cite{santra-etal-2021-hierarchical,pappagari2019hierarchical} 
operating on the original text. In addition to improving performance, the use of discrete structural cues in classification can support conversation analysis. For example, we can identify seller strategies that are more likely to lead to a successful outcome or use the sub-dialogue state sequence to summarize frequently visited states in unsuccessful interactions.

Since hand-annotation of structure can be costly and inventories vary across tasks, there is substantial interest in unsupervised learning of structure for specific task domains.
Here, the approach to structure learning involves two steps. First, we use a clustering algorithm to learn a mapping of utterance embeddings to discrete categories, which serve as an unsupervised version of DAs. 
Each conversation is then represented by the discrete sequence of cluster identifiers (IDs) associated with the sequence of utterances.  Using the collection of discretized conversations, we automatically learn the topology of a latent finite-state model over these sequences, i.e., a hidden Markov model (HMM), using a greedy state-splitting algorithm that maximizes the likelihood of the sequence data without requiring any annotations. 
The states of the HMM correspond to different sub-dialogues that may be associated with specific topics, strategies or sub-tasks. The sub-dialogue structure of a new conversation is identified by finding the most likely state sequence given that discretized utterance sequence. 

The learned structure is assessed in experiments on three conversation-level classification tasks: buyer/seller negotiation outcomes on \cb \cite{he-etal-2018-decoupling}, conversation category in the Action-based Conversations Dataset (\abcd) \cite{chen-etal-2021-action}, and client callback prediction in a private call center corpus. In each task, we find that a combination of both utterance-level category and sub-dialogue state information lead to improved performance. Further, we use automatically generated descriptions of the clusters and sub-dialogue states to provide an interpretable view of the finite-state topology and a summarized view of a conversation. Anecdotally, we find that this structure lends insights into how participant strategies (state paths) are associated with different conversation outcomes. 

The contributions of this work are as follows. First, we introduce a simple unsupervised approach to learn a hierarchical representation of
conversation structure that includes turn-level labels and sub-dialogue segmentation, accounting for participant role.  Using three conversation-level classification tasks, we demonstrate that integrating the
structural information into a state-of-the-art hierarchical transformer consistently improves performance.
Lastly, we show how the discrete representation of structure combined with automatic summarization can provide a mechanism for interpreting what the model is learning or for conversation summarization and analytics.

%% file: method2.tex
\begin{figure*}[ht!]
\centering
    \includegraphics[width=.95\linewidth]{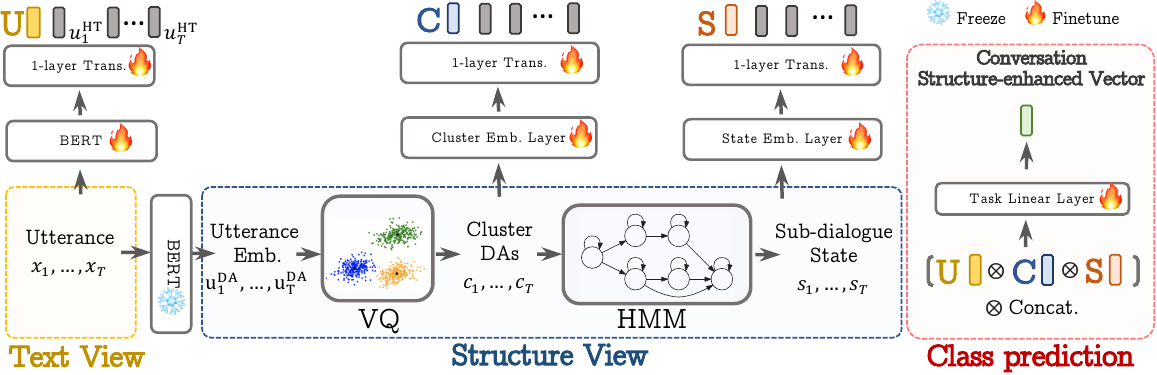}
    \caption{Overview of \ours conversation encoding. The text of each utterance text is encoded by BERT, and a 1-layer transformer further contextualizes utterance embeddings to generate the text vector $\mathbf{U}$. For structure, utterances are mapped to K-means dialogue acts (DAs), which are input to an HMM to decode sub-dialogue states. 1-layer transformers are applied to sequences of DAs and sub-dialogue states, yielding cluster vector $\mathbf{C}$ and state vector $\mathbf{S}$. The concatenation of $\mathbf{U}$, $\mathbf{C}$ and $\mathbf{S}$ is fed into a linear layer to obtain the structure-enhanced vector for the predictive task. For simplicity, Emb. and Trans. stand for embedding and transformer, respectively.}
    \label{fig:model}
\end{figure*}

\section{Method}
\label{sec:method}

As shown in  \autoref{fig:model}, \ours represents the sequence of turns in a conversation using: i) a hierarchical transformer (\ourht) operating on a turn-segmented word sequence, ii) a transformer operating on a sequence of turn-level DAs, and iii) a separate transformer operating on a sequence of sub-dialogue states derived from the DAs. 
The conversation-level vectors produced by the three transformers are concatenated and used in a final task-specific layer for conversation classification tasks. 
The \ourht alone is the state-of-the-art model for conversation-level tasks.
The DA and sub-dialogue states comprise the structural information that enhances the \ourht for improving performance of the end task. 
In addition, the discrete nature of the structure representation provides a mechanism for analyzing the conversation classes via summarization of utterances associated with the DA labels or sub-dialogue states.

\subsection{Model Components}

\paragraph{Definitions}


More formally, each dialogue consists of a sequence of words (or tokens) $X = [x_1, \ldots, x_T]$ associated with $T$ customer/agent (or seller/buyer) utterances, where $x_t$ is the subsequence of words associated with the $t$th utterance.\footnote{We use the term ''utterance'' although some conversations involve text-based interactions.}
The word sequence is decorated with three special tokens: \cls, \ptytoken and \septoken, 
where \ptytoken indicates the utterance speaker role
(\agttoken for agent/seller and \usrtoken for customer/buyer). 
The word sequence $X$ is mapped to two sequences of utterance-level embeddings $U^v = [u_1^v, \ldots, u_T^v]$, $v\in\{\ourht, \da\}$. The vector $u_t^{\ourht}$ is output from the last layer of the \ourht that is used to derive the text-based conversation-level vector $\mathbf{U}$.
The vector $u_t^{\da}$ is the output of a separate transformer, which is then mapped to a DA category $c_t$ to produce the sequence $C = [c_1, \ldots, c_{T}]$. 
The sequence $C$ is associated with a hidden subdialogue sequence that is represented using the HMM state sequence $S= [s_1, \ldots, s_{T}]$.
Additional transformers derive conversation-level vectors $\mathbf{C}$ and $\mathbf{S}$ from $C$ and $S$, respectively.
\ours enhances the conversation representation by concatenating $\mathbf{U}$, $\mathbf{C}$ and $\mathbf{S}$ together for input to a task-specific layer.

\vspace{-1mm}

\paragraph{Hierarchical Transformer}
The hierarchical transformer \cite{pappagari2019hierarchical} has been shown to be useful for classifying long documents (like customer support conversations), which exceed the length limits placed on transformer-based models due to the quadratic complexity of the self-attention module.
At a high level, two transformer blocks, a lower utterance transformer and an upper conversation transformer are stacked together for encoding dialogues.
Here, the utterance-level transformer first encodes utterances into utterance embeddings, one for each utterance.
In this case, the first contextualized token embedding as the utterance embedding, 
which corresponds to the sentence-level \cls token. 
The sequence of utterance embeddings augmented with a conversation-level \cls token are then fed as inputs to another one-layer conversation-level transformer to further contextualize the vector sequence. We use the output vector associated with the conversation-level \cls token as the conversation representation.

\vspace{-1mm}

\paragraph{Dialogue Act Sequence Module}
To obtain the DA labels, we first derive
an utterance embedding $u^{\da}_t$ by mean pooling the final layer of the BERT transformer.\footnote{We also experimented with using the \cls token, but mean pooling gave better results.} The resulting embedding is mapped to a DA class $c_t$ using a vector quantization (VQ) approach: K-means clustering is used to learn the classes, and vectors are labeled at inference time by minimizing the Euclidean distance to cluster means. 
The number of clusters is treated as a hyperparameter of the overall model.
We apply K-means clustering separately for utterances from the two different participant roles, so the DA index reflects the role.
This simple approach is motivated by prior work on unsupervised learning of DA categories \cite{BrychcinKral17}, which showed that K-means clustering gives a performance that is only slightly worse than HMM-based learning.

In linguistic analyses, a turn can contain a sequence of DAs. Our work assigns a single DA to a user turn, as in other work using unsupervised learning as well as the negotiation data set that we report results on. Since the prior work often uses ``dialogue act'' for turn-level labels, we have chosen to use the DA term here, acknowledging the abuse of terminology. For complex tasks like the call center data (and other data with real users), the turns will involve multiple dialogue acts, in which case a large number of clusters is useful.

\paragraph{Sub-Dialogue Sequence Module}
The DA sequence $C$ is input to a hidden Markov model (HMM) to derive the sub-dialogue structure.  
An HMM is a statistical model that characterizes an 
observation sequence $C$
in terms of a discrete, latent (hidden) Markov state sequence  $S$,
\[
  \begin{split}
    P(C) &= \sum_{\text{all}\ S}P(C,S) \\
    &= \sum_{\text{all}\ S}\pi(s_1) \prod_{t=1}^{T} \eta(c_{t}|s_{t}) \gamma(s_{t+1}|s_{t}),
  \end{split}
\]
where $\pi$, $\eta$, and $\gamma$ are start-state, observation, and transition distributions, respectively. $s_{T+1}$ is a dummy stopping state.
The HMM is used to decode the hidden sub-dialogue state sequence $S$, which
provides a segmentation of the conversation into  different stages or sub-tasks in problem solving or negotiation.
The HMM topology and parameters are derived using unsupervised learning as described in the next section.

\subsection{Sub-Dialogue Structure Learning}
\label{sec:topology_learning}

\begin{figure*}[ht!]
  \begin{subfigure}[t]{.32\textwidth}
    \centering
    \includegraphics[width=0.7\linewidth]{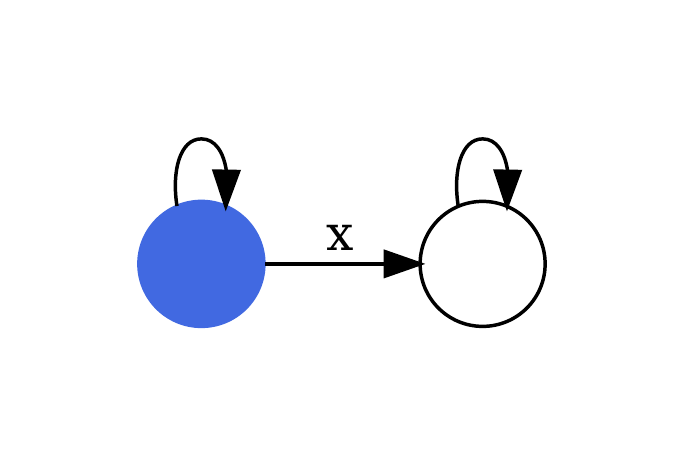}
    \caption{Before split.}
    \label{fig:before_split}
  \end{subfigure}%
  \begin{subfigure}[t]{.32\textwidth}
    \centering
    \includegraphics[width=0.73\linewidth]{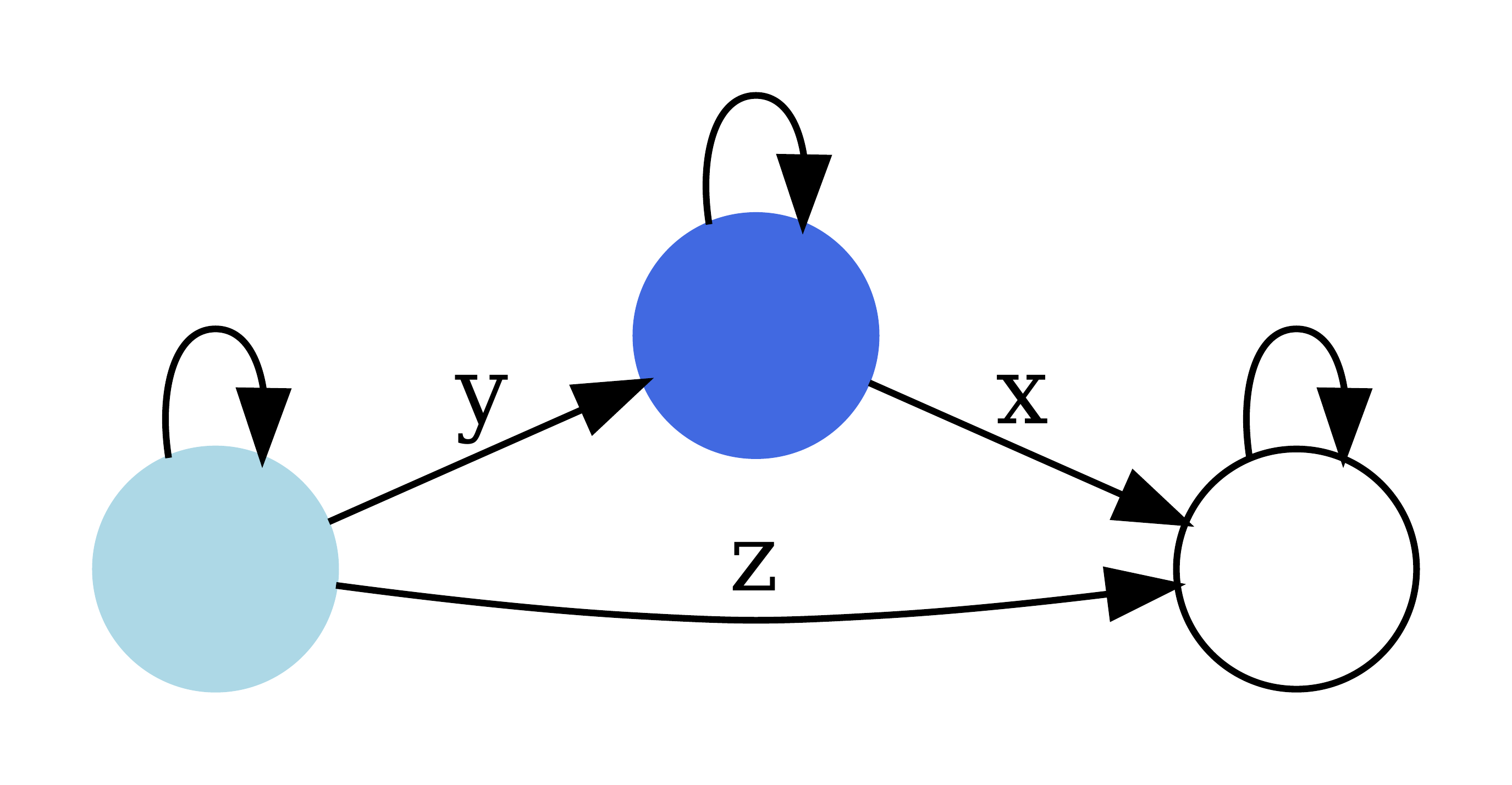}
    \caption{Temporal split.}
    \label{fig:temporal_split}
  \end{subfigure}
  \begin{subfigure}[t]{.32\textwidth}
    \centering
    \includegraphics[width=.42\linewidth]{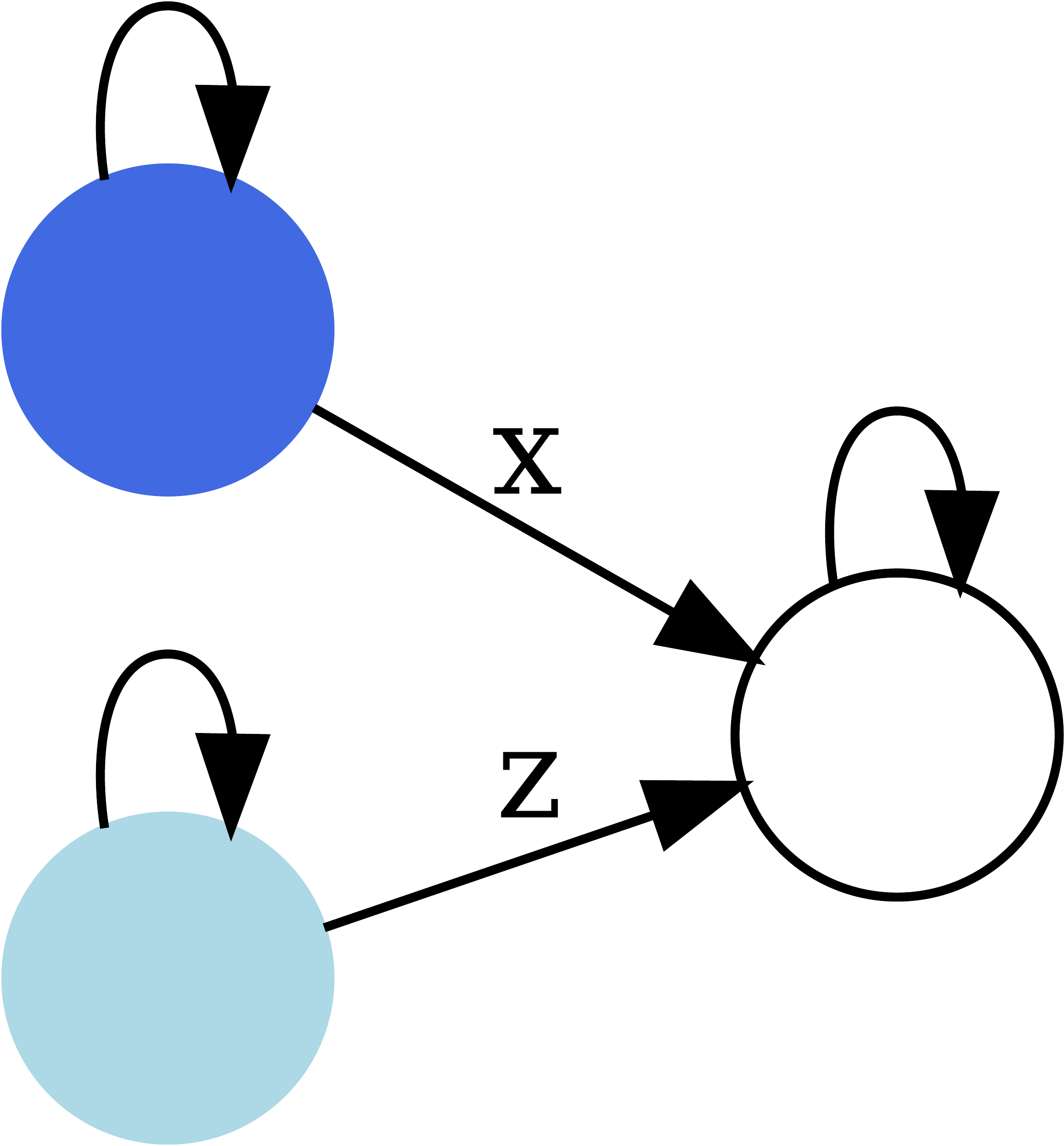}
    \caption{Contextual split.}
    \label{fig:contextual_split}
  \end{subfigure}
  \caption{The design of two split methods. The dark-blue state is chosen to be split. The light-blue state is the new state after split. Transitions to other states are omitted for simplicity.}
  \label{fig:split}
\end{figure*}

\vspace{-2mm}

Given a specified topology, inference and training algorithms for HMMs are well established \cite{Murphy:2012}; the Viterbi algorithm gives the most likely state sequence, and the Expectation-Maximization (EM) algorithm is used for parameter estimation. 
To automatically learn the HMM topology, we apply
a greedy state splitting algorithm \cite{ostendorf1997hmm}, which learns a left-to-right topology by constraining states to inherit the transition constraints of their parent. 
The standard objective is maximum likelihood of the DA sequence, which is unsupervised with respect to the conversation-level task.

Topology learning is outlined in Algorithm~\ref{alg:state_splitting}. The initial model has a 3-state left-to-right topology, initialized (assuming 70\% of the conversation is associated with the middle state) and then iteratively trained until the improvement is lower than a fixed threshold or the iteration count exceeds some number. At each iteration, the state with the highest entropy of emission probability is chosen to be split. The topology can change into two new configurations corresponding to temporal and contextual splits (Figure~\ref{fig:split}). The EM algorithm is applied again on each configuration and the topology that leads to the higher likelihood is chosen. We iteratively conduct the splitting until the total number of states reaches the desired value (a hyperparameter). The HMMs of \citet{ostendorf1997hmm} used continuous observation distributions.  The splitting approach described below was designed for discrete distributions. 


\alglanguage{pascal}
  \begin{algorithm}
  \small
  \begin{algorithmic}[1]
    \State $n$: number of split. $\tau_i$: topology after $i$ split. 
    \State Initialization: Run the EM algorithm on 3-state initial topology $\tau_{0}$.
    \For{i = 1}{n}
    \Begin
        \State \begin{varwidth}[t]{0.9\linewidth}
         Select state $s\in \tau_{i-1}$ to split based on max entropy of observation distribution $\eta_{i-1}(c|s)$.
        \end{varwidth}
        \State Apply temporal split and get new topology $\tau_{i,t}$.
        \State Apply contextual split and get new topology $\tau_{i,c}$.
        \State Run the EM algorithm on $\tau_{i,t}$ and $\tau_{i,c}$.
        \State Select the topology with higher likelihood as $\tau_{i}$.
    \End
  \end{algorithmic}
  \caption{Topology Learning Algorithm}
  \label{alg:state_splitting}
\end{algorithm}


\vspace{-3mm}

\paragraph{Temporal split}
The temporal split provides more detailed sequential structure along a path.
Figure~\ref{fig:temporal_split} shows the result of a temporal split on the selected state (dark-blue) in Figure~\ref{fig:before_split}. The light-blue node is the new child state that inherits all incoming and outgoing edges and the transition probabilities of the dark-blue state except $y$ and $z$. Edges $y$ and $z$ are initialized to $p_x/2$ where 
$p_x$ is the probability of the original edge $x$ of dark-blue state. 
The old incoming edges of dark-blue state are removed and outgoing edges are preserved.

\vspace{-1mm}

\paragraph{Contextual split}
The contextual split allows for alternate sub-dialogue paths. 
\autoref{fig:contextual_split} illustrates the contextual split applied on the dark-blue state. The light-blue state inherits everything but the observation distribution of dark-blue state. With the aim of modeling different types of paths, when copying the observation probabilities to the light-blue state, we omit the top emission probability of the dark-blue node and set it to 0 and normalize the rest of probabilities. In terms of the transition probabilities, the light-blue state inherits all from the dark-blue one; $p_x = p_z$ where $p_x$ and $p_z$ are the transition probabilities of edges $x$ and $z$.

\vspace{-1mm}

\subsection{Pre-Training and End-Task Training}
\label{ssec:our_model_for_prediction}

Both for initializing the \ourht and for deriving the DAs, we use the transformer-based BERT model \cite{devlin-etal-2019-bert} for encoding individual utterances $u_t$, pre-trained using masked language modeling 
and next-sentence prediction.
Due to the style differences of dialogue data vs.\ written text, we apply domain-adaptive pretraining (DAPT) \cite{gururangan-etal-2020-dont} to adapt BERT for dialogue applications.
As shown later (\autoref{sec:exp}), adapting BERT with DAPT provides substantial improvement in terms of predictive power as well as optimization stability.


For the \ourht alone, supervised training involves learning the weights of the final task-level linear layer, the utterance-level transformer, and the word-level transformer.

For {\ours}, supervised training involves learning the weights of the cluster- and state-level transformers, in addition to all updates associated with the \ourht component described above. 
The cluster sequences are obtained using the word-level transformer with DAPT and the associated cluster mapping obtained from unsupervised learning, i.e., without task-level finetuning.  Similarly, there are no task-level supervision updates to the parameters associated with the HMM that is used to derive the state sequence.

%% file: experiment.tex
\section{Experiment}
\label{sec:exp}

\subsection{Datasets and Evaluation Metrics.}

We use three datasets with conversation-level classification tasks to evaluate our model. The detailed statistics of the datasets are shown in \autoref{sec:appendix_data}.

\cb \cite{he-etal-2018-decoupling} is a public negotiation dataset where buyers and sellers negotiate the prices of items on sale. In each conversation, the buyer has a target price in their mind and attempts to reach an agreement with the seller. Following previous work \cite{Zhou2020Augmenting,joshi2021dialograph}, we use the same list of 14 handcrafted utterance DAs and the 5-class sale-to-list price ratio labels provided in their code base. The 14 handcrafted utterance DAs are used as comparison to evaluate if our unsupervised version of DAs is learning good representations. Classification of sale-to-list price ratio is used as the downstream task, with accuracy as the evaluation criterion. 

\abcd \cite{chen-etal-2021-action} is a public customer support dataset that is introduced to study customer service dialogues.
In each conversation, an agent follows guidelines to help a customer solve their issue. 
Conversations are categorized with flows and subflows. Flows are broad categories, such as shipping issue, account access, or purchase dispute. Subflows comprise 96 fine-grained labels, for example, shipping status question, recover password, or invalid promotion code. Each conversation is annotated with a flow and a subflow. We use classification of the subflows as our conversation-level task.  Macro and micro F1 scores are used to reflect the performance of imbalanced subflow classes.

\callcenter is a private collection of customer service conversations. Phone calls are automatically transcribed and private user information is anonymized.
Conversations are annotated with a binary indicator as to whether or not there will be a callback within two days. (Such callbacks are an indicator that the problem was not solved in the call.)
For the task of callback prediction, we measure area under the ROC curve (ROC AUC).

\subsection{Implementation Details}
\paragraph{Experimental Setup.} We develop our K-means and HMMs using the packages Faiss \cite{johnson2019billion} and Pomegranate \cite{schreiber2018pomegranate}. The number of DAs and the size of the HMM state space are chosen separately for each dataset based on development set performance. We initialize and finetune our experiments based on uncased base model of BERT downloaded from HuggingFace \cite{wolf-etal-2020-transformers}. We DAPT with dynamic whole-word masking (WWM) on 128-token segments for each dataset. During finetuning, the learning rate and warm-up steps are $1 \times 10^{-5}$ and $0.1$ epoch, respectively. Models are selected by the best score on the development set for each dataset.  Further hyperparameter details are in \autoref{sec:appendix_exp}.

\input{tables/craigslist}
\vspace{-5mm}
\input{tables/abcd_tmobile_test}
\input{tables/ablation}

\vspace{-1mm}
\subsection{Comparison Systems}
We use the hierarchical transformer (\ourht) as a baseline for all datasets in comparison to \ours.
For \cb, we also include three additional baselines from two works \cite{Zhou2020Augmenting,joshi2021dialograph} that employ the DAs extracted by heuristic methods; our systems use K-means to obtain primitive DAs.


\paragraph{FST-enhanced hierarchical encoder-decoder model (FeHED).} FeHED \cite{Zhou2020Augmenting}
uses an RNN-based sequence-to-sequence model with finite-state transducers for encoding sequences of strategies and DAs.

\vspace{-1mm}

\paragraph{Hierarchical encoder-decoder (HED) + RNN or transformer.}
HED encodes dialogue utterances with a transformer (initialized from pretrained BERT), and the decoder generates the next response. An RNN or transformer encodes strategies and DAs. HED + RNN is based on the dialogue manager of \citet{he-etal-2018-decoupling}; \citet{joshi2021dialograph} replace the RNN with a transformer.

\vspace{-1mm}

\paragraph{\dialograph.} \cite{joshi2021dialograph}. The state-of-the-art HED-based model  on \cb dataset leverages graph attention networks (GAT; \citealp{velivckovic2017graph}) to encode strategies and DAs.

\subsection{Prediction Results}
\paragraph{Performance on Negotiation Dialogues.}
\autoref{tab:craigslist} reports the results of different systems on the test set of \cb dataset. All models are based on the BERT-base model.
\ourht with only text outperforms the state-of-the-art \dialograph which leverages a graph-based representation of conversation structure.
This verifies our hypothesis that DAPT with target data indeed improves BERT for dialogue tasks.
Compared with \ourht, \ours achieves better prediction accuracy and smaller variance, which suggests that integrating the structure view helps stabilize training with different random seeds.
\ours provides a 24.5\% relative gain in accuracy over \dialograph, setting a new state of the art. 
This further validates the advantage of our learned conversation structure for a  predictive task. 

\vspace{-1mm}

\paragraph{Performance on Customer Support Domain.}
Similar to the results on the negotiation dialogue domain, \autoref{tab:abcd_tmobile_test} shows that conversation structure effectively enhances the performance in the customer service domain, \abcd and \callcenter. 
\vspace{-1mm}

\paragraph{Ablation.} \autoref{tab:ablation} reports the results of ablating different components of \ours on the validation sets of all datasets.
The first rows show that DAPT is useful on all tasks particularly for \abcd with its skewed class distribution.  We also observe that \ours consistently achieves the best performance over all tasks. The cluster-based DA sequence provides more information than the sub-dialogue states, but incorporating
all three views together leads to the best performance. Statistical significance is tested using bootstrap resampling \cite{Efron1993AnIT, berg-kirkpatrick-etal-2012-empirical}. 

Prior work \cite{Zhou2020Augmenting, joshi2021dialograph} on \cb use domain knowledge in rule-based annotation of DAs. To assess the use of K-means clusters for learning DAs, we also trained an HMM using the provided DAs. The resulting model obtained 66.5\% accuracy on the test data, which is not significantly different the 66.1\% results obtained using K-means (cf.~Table~\ref{tab:craigslist}).



%% file: tables/craigslist.tex
\begin{table}[ht!]
\small
\centering

\resizebox{0.62\linewidth}{!}{
  \begin{tabular}{lcc}
    \toprule
    Model                      & \% Acc.                   \\ \midrule
    FeHED                      & 42.3                   \\
    HED + RNN                  & 47.9                   \\
    HED + transformer          & 53.7                   \\
    \dialograph                & 53.1                   \\ \midrule
    \ourht                     & 54.1{\tiny $\pm$ 2.4}  \\
    \ours                      & \bf{66.1}{\tiny $\pm$ 1.0}  \\\bottomrule
    \end{tabular}
}
\caption{Results on the test set of \cb in accuracy. For models studied in this paper (lower part), the median number is reported with standard deviation calculated based on 15 random runs. 
}
\label{tab:craigslist}
\end{table}

%% file: tables/abcd_tmobile_test.tex
\begin{table}[ht]
\centering
\small
\resizebox{\linewidth}{!}{
\begin{tabular}{lcccc}
  \toprule
           & \multicolumn{3}{c}{\abcd} & \callcenter \\\cmidrule{1-5} 
           & \multicolumn{3}{c}{F1}    &          \\
  Model    & Micro & Macro & Weighted  & ROC AUC  \\\midrule
  \ourht   & 52.2  & 25.4  & 45.7      & 69.6     \\
  \ours    & \bf62.8  & \bf39.1  & \bf59.9      & \bf71.3     \\\bottomrule
  \end{tabular}
}
\caption{Results on the test sets of \abcd and \callcenter datasets.}
\label{tab:abcd_tmobile_test}
\vspace{-1mm}
\end{table}

%% file: tables/ablation.tex
\begin{table*}[ht!]
\centering
\small
\resizebox{0.8\textwidth}{!}{
\begin{tabular}{lccccc}
  \toprule
                            & \cb      & \multicolumn{3}{c}{\abcd} & \callcenter \\\cmidrule{2-6} 
                            &          & \multicolumn{3}{c}{F1}    &          \\
  Model                     & Accuracy & Micro & Macro & Weighted  & ROC AUC  \\\midrule
  \ourht w/o DAPT           & 48.0     & 15.4  &  4.2  &  9.4      & 68.4     \\
  \ourht                    & 50.3     & 52.2  & 26.9  & 46.3      & 71.2     \\\midrule
  \ours (cluster only)      & 60.2     & 59.8  & 35.3  & 55.7      & 72.2     \\
  \ours (state only)        & 51.7     & 58.8  & 32.8  & 54.1      & 72.1     \\
  \ours                     & \bf61.3     & \bf62.6  & \bf38.6  & \bf59.5      & \bf72.8     \\\bottomrule
  \end{tabular}
}
\caption{Ablation on the development sets of \cb, \abcd and \callcenter datasets. All models with structure are statistically better than \ourht.  \ours is better ($p < 0.01$) than the cluster-only alternative except for the \callcenter.}
\label{tab:ablation}
\end{table*}

%% file: analysis.tex
\vspace{-1mm}
\section{Interpretation and Analysis}
\label{sec:analysis}
In this section, we leverage automatic summarization of clusters and states to derive insights into the learned conversation structure, both for interpretability of the model and for applications such as conversation analytics and summarization. 
As an example, we analyze fine-grained components from the learned topology, i.e., most frequent paths and individual state n-grams, to investigate their associations with different dialogue characteristics.



We apply graph-based unsupervised summarization \cite{boudin-morin-2013-keyphrase,shang-etal-2018-unsupervised} 
over utterances in each state (decoupling participant roles) and in each cluster.
On \cb and \abcd, this leads to more than 3$\times$ reduction in conversation length.

\begin{figure*}[ht!]
\centering
  \includegraphics[width=0.95\linewidth]{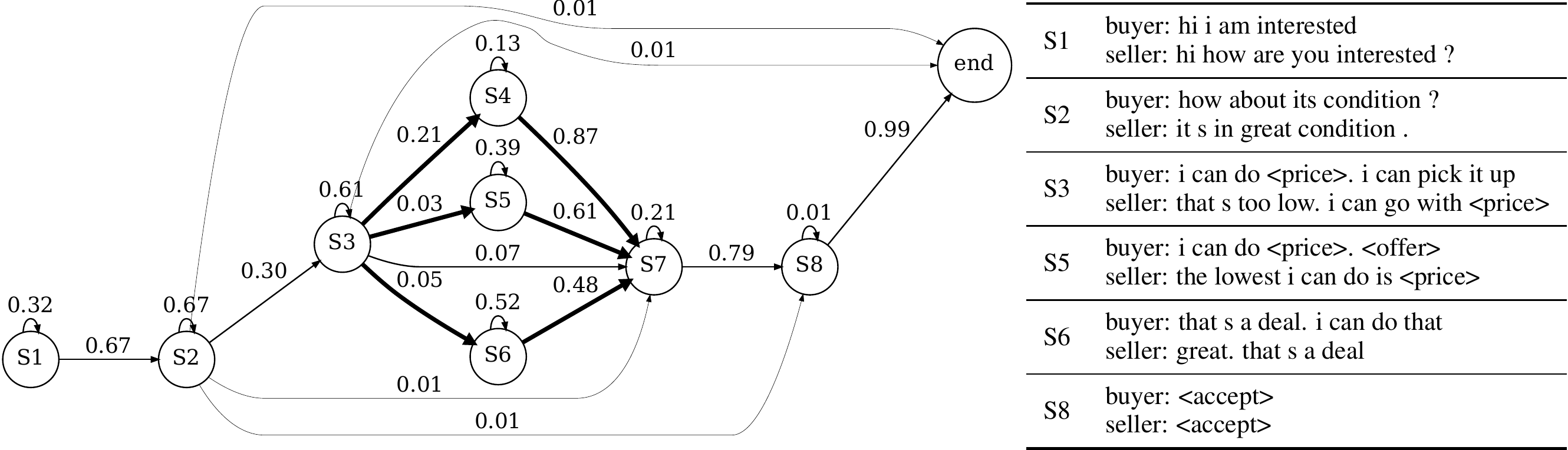}
  \caption{The 8-state topology on \cb dataset. The thicker edges indicate higher levels of negotiation success; in contrast, the thinner edges represent lower levels. Due to space limitations, only 6 state summaries are shown. The detailed topology with all cluster and state summaries is in \autoref{sec:appendix_topology}.}
  \label{fig:cb_topology}
\end{figure*}

\autoref{fig:cb_topology} shows the 8-state topology of \cb with selected state summaries.
Based on the summaries, it is easy to see that \texttt{S1} and \texttt{S8} capture opening and closing DAs, respectively, while \texttt{S5} and \texttt{S6} correspond to different negotiation strategies.
We also find that conversations with shorter paths are likely to involve a less experienced seller or lower buyer interest, e.g., $92\%$ conversations with path \texttt{S1-S2-S8} lead to under listing sells.
On the other hand, sellers that say offers are too low are more likely get better prices, e.g., $91\%$ conversations with path \texttt{S1-S2-S3-S5-S7-S8}.

\input{tables/abcd_summarization_example}

\paragraph{\abcd}
\autoref{tab:abcd_example} shows an example with both cluster and state summaries.
Based on the cluster summaries, we see that K-means learns typical DAs associated with customer service, e.g., information requests from the agent and the corresponding customer replies (\texttt{<name>, <email>, <address>}).
States correspond to sub-dialogues where the agent follows certain protocols in resolving a sub-task (e.g., verifying account information).
Alignment of flow labels with the most frequent paths through the HMM topology shows that paths are highly indicative of the corresponding dialogue flow.
The high confusions are among certain flows, such as \texttt{storewide\_query} and \texttt{single\_item\_query}, which one would expect to have have similar DAs.

%% file: tables/abcd_summarization_example.tex
\begin{table*}[ht!]
\centering
\small
\resizebox{.95\textwidth}{!}{

  \begin{tabular}{rl|l|l}
  \toprule
  \multicolumn{1}{c}{Party}    &  \multicolumn{1}{c}{Utterance}  &  \multicolumn{1}{c}{Cluster Summary} & \multicolumn{1}{c}{State Summary}                                            \\
  \midrule
Agent    & Welcome to AcmeBrands! How can I help you?  &   How can I help you?     & \multirow{3}{*}{\shortstack[l]{A: How can I help you today?\\C: I want to check my order.}}\\
Customer & Hello, I would like to change my shipping details & \multirow{2}{*}{I want to check my shipping} & \\
& as they have changed recently due to a move & & \\
\hline
Agent & I would be happy to help you with that  & I can help you with that & \multirow{12}{*}{\shortstack[l]{A: Can I have your account/order?\\C: My account/order is \_\_\_\_\_\_\_}}\\
Agent & Is there an outstanding order? &  How long have you been waiting? & \\
Agent & Or is this just an update to your account? & I have pulled up your account. & \\
Customer & Yes my order id is 4870952797 & My order/account ID is \_\_\_\_\_\_\_ & \\
Agent & What is your name please? & What is your name? & \\
Customer & Crystal Minh  & <name> & \\
Agent & What is the shipping status of the order? & What is the shipping status? & \\
Customer & In Transit & In store/ In transit & \\
Agent & Next I need to validate your purchase.  & \multirow{2}{*}{I need your name} & \\
& I will need your username and email. & & \\
Customer & cminh948, cminh948@email.com & <email> & \\
Agent & Thank you & Thank you&\\
\hline
Agent & and the new address please? & Can you tell me \_\_\_\_\_\_? & \multirow{3}{*}{\shortstack[l]{A: Can I have the address?\\C: My address is \_\_\_\_\_\_\_}}\\
Customer & 9756 Primrose Street Newark, MI 85971  & <address> & \\
Agent & All taken care of! & Your order has been updated & \\
\hline
Agent & Is there anything else today? & Anything else? & \multirow{3}{*}{\shortstack[l]{A: Anything else I can help?\\C: That's all. Thank you.}} \\
Customer & Thank you that is all  & That's all. Thank you. & \\
Agent  & Have a great one!  & Have a good one! & \\
  \bottomrule
  \end{tabular}
}
\caption{An example of \abcd with cluster and state summaries. A and C stand for agent and customer, respectively.}
\label{tab:abcd_example}
\end{table*}

%% file: related2.tex
\section{Related Work}
\label{sec:related_work}
\vspace{-2mm}

HMMs have been leveraged for learning structure in language for many years, such as in early work on inducing word-level part-of-speech tags (\citealp{merialdo-1994-tagging}).  Accordingly, most work on unsupervised learning of both DAs and conversation structure leverages HMMs.

\paragraph{Unsupervised Learning of Dialogue Acts.}

Since dialogue act recognition can be thought of as a sentence-level tagging task, initial work on unsupervised learning of DAs was similar to word tagging, involving some use of language models or fully-connected HMMs to account for sequential dependency of labels.
\citet{ritter-etal-2010-unsupervised} use an HMM with a factored state space with a topic model to decouple speech act from topic characteristics.
The observation model $\eta$ in the HMM is a bag of words (unigram) model. 
The approach was later extended by incorporating speaker information \cite{Joty+11,paul-2012-mixed}.
\citet{BrychcinKral17} further extend this work with a Gaussian mixture observation model (GMM) where the utterance representation is the average of GloVe word embeddings. They compare the results to a simple K-means clustering, which is not as effective as the HMM but gives similar results to the method proposed by \citet{ritter-etal-2010-unsupervised} when applied to the Switchboard corpus.
Hierarchical clustering of delexicalized utterance embeddings is used by \citet{GUNASEKARA2019}, who use domain knowledge in preprocessing to identify phrases such as ``Indian food'' as ``CUISINE\_TYPE,'' for example.
Our work on utterance categorization is similar to the K-means approach in \citet{BrychcinKral17}, but we use more recent transformer-based utterance embeddings. 

\paragraph{Unsupervised Learning of Dialogue Structure.}

Task- or goal-oriented conversations typically have structure above the level of the sentence in that a sequence of turns are associated with a common function. In more complex conversations, the structure can be hierarchical, with tasks and sub-tasks.  \citet{Bangalore+08} used a parsing model to automatically recognize dialogue acts and segment a conversation into sub-tasks, leveraging hand-annotations of both DAs and sub-tasks. Since sub-task structure varies depending on the task and there is little hand-annotated data, most work has focused on unsupervised approaches with a flat segmentation.
Note that the problem of unsupervised learning here involves jointly recognizing sub-dialogue segment boundaries, learning an inventory of sub-dialogue types, and learning (or constraining) the sequential structure of these types.

Early work on unsupervised learning used fully-connected HMMs to identify structure in documents \cite{barzilay-lee-2004-catching} for extractive summarization and information ordering. The observation model was based on word bigrams with the aim of capturing topic coherent segment. A similar idea is applied to task-oriented dialogues using latent Dirichlet allocation for the observation model \cite{zhai-williams-2014-discovering}.

Studies that leverage constrained left-to-right HMM technologies include \cite{althoff-etal-2016-large}, which aimed to learn stages/strategies of counselors in mental health counseling, and   \cite{ramanath-etal-2014-unsupervised}, which used a hidden semi-Markov model for unsupervised alignment of privacy policy documents. Both used unigram observation models. HMM-based conversation stages are combined with a topic-based segmentation by \citet{chen-yang-2020-multi} for dialogue summarization.
The use of unigram and bigram word models emphasizes topic in segmenting conversations.
Our work differs in that the automatically learned speech acts are observations of the HMM, since word distributions are captured by the \ourht. 

Most similar to our work is \cite{Zhou2020Augmenting}, which uses two finite state transducers (FSTs) to map a sequence of dialogue acts (or strategies) to a sequence of state embeddings, which are then integrated into a hierarchical encoder-decoder model for prediction of the next strategy in a negotiation dialogue. The FSTs are analogous to our HMM, but the inputs are based on learning from hand-labeled strategies and rule-based dialogue acts.

There are other approaches to modeling conversation structure that do not rely on HMMs. \dialograph \cite{joshi2021dialograph} uses a graph attention network to encode discrete DA and strategy label sequences. 
A variational recurrent neural network is used to model structure by \citet{shi-etal-2019-unsupervised}. These approaches are less amenable to the interpretation methods used in our work.

Two key differences in our approach compared to all these studies are: i) the use of HMM topology learning via successive state splitting, and ii) the integration of structural information using a multi-stream neural sequence model.

%% file: conclusion.tex
\section{Conclusion}

In summary, this work combines two simple approaches for unsupervised learning on top of embedded utterance representations (K-means clustering and HMM topology design) to derive a hierarchical representation of conversation structure, which is useful to enhance a hierarchical transformer in three conversation-level classification tasks.
The K-means clusters are intended to approximate DAs
and the HMM is intended to learn sub-dialogue structure. Unlike prior work in this area, the sub-dialogues build on DA sequences rather than unigram/bigram statistics, and the HMM incorporates forward-moving dialogue flow constraints in topology learning, with the goal of capturing sub-dialogue function.




%% file: acknowledgement.tex
\section*{Acknowledgments}
We thank Mourad Heddaya for exploring preliminary experiments when he was at the University of Washington, and Chih-Ning (Sonia) Ho for help with experiments and technical discussions. We also thank all members of the TIAL lab and NLP groups at the University of Washington who provided suggestions and insights into this work. 

%% file: limitation.tex
\section*{Limitations}


First, our experiments explore only two types of dialogues (negotiation and customer support) with conversation-level tasks (identifying the topic or assessing some measure of conversation success).
Although \ours shows promising results, it requires further exploration with other types of conversations (e.g.\ information gathering, tutoring), including more examples of spoken interactions, as well as extending \ours to multi-party discussions.
In addition, it would be of interest to assess the utility of automatically learned structure for other types of tasks, such as call center analytics or state tracking to support dialogue management or online agent support.

Second, we use K-means and HMMs for deriving the conversation structure, both of which 
require dataset-specific hyperparameters that are unlikely to transfer well to new datasets.
Additionally, we only study a late fusion strategy for combining discrete structure and text-based representations.
A more tightly integrated approach might be more effective.
For example, our K-means DA is based on a single utterance; however, sequence models have been important for past work on unsupervised learning of DAs.
Future work could leverage sequential DA dependencies in joint DA and sub-dialogue structure learning or explore continuous DA-like representations, as in \cite{cheng-etal-2019-dynamic}.

%% file: ethical.tex
\section*{Ethical Considerations}
The automatic learning of conversation structure is dependent on having data that is matched to the task of interest. A potential challenge is that biases in the data could result in some conversation strategies not being well represented.
The summarization approach provides interpretability of the model, but imperfect summarizations could lead to incorrect interpretations.

%% file: appendix.tex
\section{Experimental Setup Details}
\label{sec:appendix_exp}
We pretrain and finetune on BERT \cite{devlin-etal-2019-bert} downloaded from Huggingface Transformers \cite{wolf-etal-2020-transformers}\footnote{\url{https://github.com/huggingface/transformers}} and use the uncased base model of BERT in most of our experiments. To feed lengthy conversations to the model, we employ gradient checkpoint and DeepSpeed \cite{rasley2020deepspeed}, a deep learning optimization library, to reduce GPU memory usage and accelerate the training process.

The details of the model hyperparameters are as follows. 1-layer and 2-head transformers with 300 hidden size are applied to encode sequences of utterance-level embeddings in text view and sequences of clusters and states in structure view. Thus, the total number of parameters of our best system \ours, including base model of BERT and 3 one-layer transformers, is about 113M. For in-domain adaptation pretraining (DAPT), we use $5 \times 10^{-5}$ as learning rate and 5000 steps for \cb and \abcd and 30000 steps for \callcenter. $0.1$ epochs are used as warm-up steps with linear learning rate decay. Gradient accumulation and PyTorch \cite{NEURIPS2019_9015} distributed data parallel GPU training are applied to achieve the equivalent training batch size 4096. For finetuning, we set $1 \times 10^{-5}$ as the learning rates, 4 epochs in total and $0.1$ epochs for warm-up steps with linear decay. The equivalent training batch size is 16 during finetuning. Besides, the layer-wise learning rate decay is utilized to stabilize the training results; the rates are from $0.7, 0.8, 0.9$ and the $0.9$ leads to the best performance. For the rest of the training hyperparameters, we follow the default values in HuggingFace's training script.

For K-means, we use Faiss \cite{johnson2019billion}\footnote{\url{https://github.com/facebookresearch/faiss}} with GPU to speed up clustering process for large private corpus. For HMMs, we develop our splitting algorithm via Pomegranate \cite{schreiber2018pomegranate},\footnote{\url{https://github.com/jmschrei/pomegranate}} a Python package that implements fast and flexible probabilistic models, to build our topology learning algorithm. The predefined numbers of clusters vary for different datasets. To compare with handcrafted DAs provided in \cb, we define number of clusters $k = 14$ for each party. For customer service domain, we set $k=60$ for \abcd and $k=120$ for \callcenter. For all datasets, we try the number of states from 5 to 20 and find the best numbers of states are 8, 12, and 12 for \cb, \abcd, and \callcenter, respectively. Each training run takes at most 2 hours on 2 Nvidia GeForce RTX 2080Ti GPUs for \cb and \abcd and 54 hours on 8 GPUs on \callcenter. All models are saved based on the best performance on the development sets. For each experiment on \cb and \abcd, we conduct $15$ random runs and report the median and standard deviation. Due to the computation limitations and the size of corpus, we only conduct a single run for \callcenter for each experiment setting. The total number of GPU hours for all experiments, including different runs with random seeds, is 1536 hours approximately.

\input{tables/dataset_stat}
\section{Dataset Details}
\label{sec:appendix_data}
We follow all original data preprocessing scripts for \cb\footnote{\url{https://github.com/rishabhjoshi/DialoGraph_ICLR21}.} and \abcd.\footnote{\url{https://github.com/asappresearch/abcd}.} For the private collection of customer service conversations, \callcenter, all private user information is anonymized. The data statistics are summarized in \autoref{tab:data_stat} and \autoref{tab:train_dev_test_split}.

\section{Topology with Summaries}
\begin{sidewaysfigure*}
    \includegraphics[width=\linewidth]{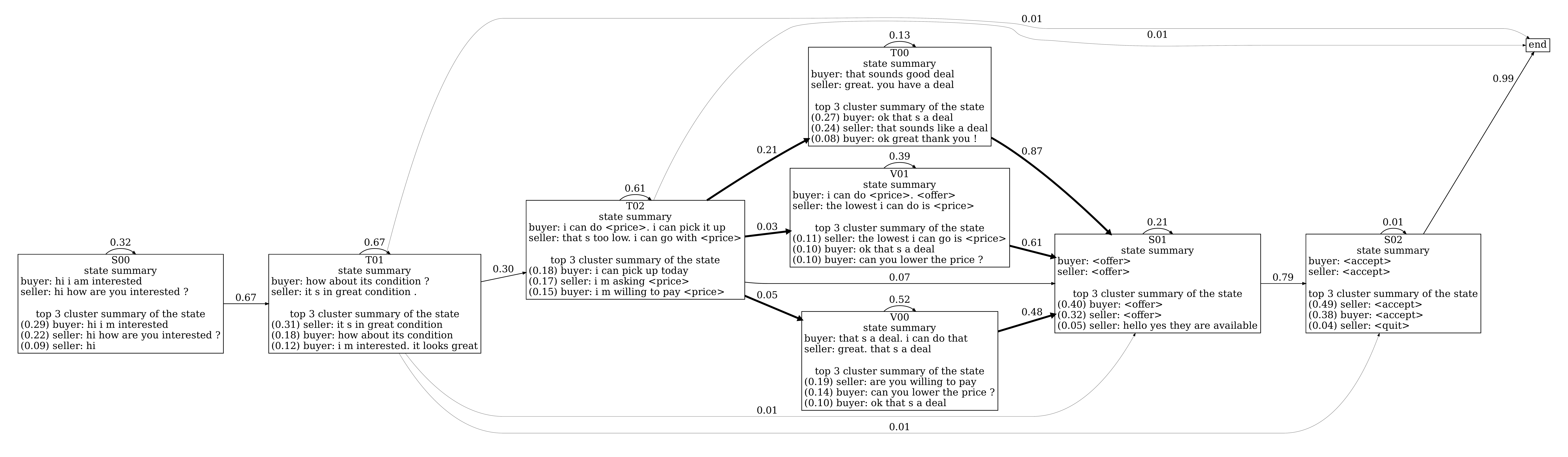}
    \caption{The 8-state full topology with cluster and sub-dialogue state summaries on \cb dataset. The thicker edges represent higher levels of negotiation success.}
    \label{fig:cb_full_topology}
\end{sidewaysfigure*}

\autoref{fig:cb_full_topology} shows the detailed topology with both cluster and sub-dialogue state summaries. For each sub-dialogue state, we add the cluster summaries with top 3 emission probabilites and the sub-dialogue state summaries for the buyer and the seller. The 
thickness of edges indicates the levels of negotiation success and the edges with probabilities lower than $0.01$ are pruned for simplicity.

\section{License of Artifacts}
\label{sec:appendix_license}
The license of code for \citet{wolf-etal-2020-transformers} and \citet{schreiber2018pomegranate} are Apache license version 2.0. The license of code for \citet{joshi2021dialograph}, \citet{rasley2020deepspeed}, and \citet{chen-etal-2021-action} are MIT License. The terms for use of our artifacts will be included in our released package.

\label{sec:appendix_topology}

%
%

%% file: tables/dataset_stat.tex
%

\begin{table}[ht]
\resizebox{\linewidth}{!}{
  \begin{tabular}{rccc}
  \toprule
                       & \begin{tabular}[c]{@{}l@{}}\textsc{Craigslist}\\ \textsc{Bargain}\end{tabular} & \abcd  & \begin{tabular}[c]{@{}l@{}}\textsc{Call}\\ \textsc{Center}\end{tabular} \\\midrule 
\# dialogues           & 6682   & 10042   & 949410       \\
  \# turns / dialogue  & 9.2                                                          & 22.1  & 71.6                                                  \\
  \# tokens / turn     & 15.5                                                         & 9.2  & 16.3                                                  \\
  \# tokens / dialogue & 142.6                                                        & 202.5 & 1167.1 \\\bottomrule
  \end{tabular}
}
\caption{Data statistics of the datasets.}
\label{tab:data_stat}
\end{table}

\begin{table}[ht]
\resizebox{\linewidth}{!}{
  \begin{tabular}{rccc}
  \toprule
                       & \begin{tabular}[c]{@{}l@{}}\textsc{Craigslist}\\ \textsc{Bargain}\end{tabular} & \abcd  & \begin{tabular}[c]{@{}l@{}}\textsc{Call}\\ \textsc{Center}\end{tabular} \\\midrule 
train set \#  dialogues         & 4828   & 8034   & 711310       \\
dev.~set \# dialogues  & 561                                                          & 1004  & 95540                                                  \\
test set \# dialogues  & 567                                                         & 1004  & 142560                                                \\
\bottomrule
  \end{tabular}
}
\caption{Train/dev./test split of datasets}
\label{tab:train_dev_test_split}
\end{table}

%% file: acl_latex.bbl
\begin{thebibliography}{39}
\expandafter\ifx\csname natexlab\endcsname\relax\def\natexlab#1{#1}\fi

\bibitem[{Althoff et~al.(2016)Althoff, Clark, and
  Leskovec}]{althoff-etal-2016-large}
Tim Althoff, Kevin Clark, and Jure Leskovec. 2016.
\newblock \href {https://doi.org/10.1162/tacl_a_00111} {Large-scale analysis of
  counseling conversations: An application of natural language processing to
  mental health}.
\newblock \emph{Transactions of the Association for Computational Linguistics},
  4:463--476.

\bibitem[{Bangalore et~al.(2008)Bangalore, Di~Fabbrizio, and
  Stent}]{Bangalore+08}
Srinivas Bangalore, Giuseppe Di~Fabbrizio, and Amanda Stent. 2008.
\newblock Learning the structure of task-driven human–human dialogs.
\newblock \emph{IEEE Transactions on Audio, Speech, and Language Processing},
  16(7):1249--1259.

\bibitem[{Barzilay and Lee(2004)}]{barzilay-lee-2004-catching}
Regina Barzilay and Lillian Lee. 2004.
\newblock \href {https://aclanthology.org/N04-1015} {Catching the drift:
  Probabilistic content models, with applications to generation and
  summarization}.
\newblock In \emph{Proceedings of the Human Language Technology Conference of
  the North {A}merican Chapter of the Association for Computational
  Linguistics: {HLT}-{NAACL} 2004}, pages 113--120, Boston, Massachusetts, USA.
  Association for Computational Linguistics.

\bibitem[{Berg-Kirkpatrick et~al.(2012)Berg-Kirkpatrick, Burkett, and
  Klein}]{berg-kirkpatrick-etal-2012-empirical}
Taylor Berg-Kirkpatrick, David Burkett, and Dan Klein. 2012.
\newblock \href {https://aclanthology.org/D12-1091} {An empirical investigation
  of statistical significance in {NLP}}.
\newblock In \emph{Proceedings of the 2012 Joint Conference on Empirical
  Methods in Natural Language Processing and Computational Natural Language
  Learning}, pages 995--1005, Jeju Island, Korea. Association for Computational
  Linguistics.

\bibitem[{Boudin and Morin(2013)}]{boudin-morin-2013-keyphrase}
Florian Boudin and Emmanuel Morin. 2013.
\newblock \href {https://aclanthology.org/N13-1030} {Keyphrase extraction for
  n-best reranking in multi-sentence compression}.
\newblock In \emph{Proceedings of the 2013 Conference of the North {A}merican
  Chapter of the Association for Computational Linguistics: Human Language
  Technologies}, pages 298--305, Atlanta, Georgia. Association for
  Computational Linguistics.

\bibitem[{Brychcin and Kral(2017)}]{BrychcinKral17}
T.~Brychcin and P.~Kral. 2017.
\newblock Unsupervised dialogue act induction using gaussian mixtures.
\newblock In \emph{Prof.\ EMNLP}, volume~2, page 485–490.

\bibitem[{Chen et~al.(2021)Chen, Chen, Yang, Lin, and
  Yu}]{chen-etal-2021-action}
Derek Chen, Howard Chen, Yi~Yang, Alexander Lin, and Zhou Yu. 2021.
\newblock \href {https://doi.org/10.18653/v1/2021.naacl-main.239} {Action-based
  conversations dataset: A corpus for building more in-depth task-oriented
  dialogue systems}.
\newblock In \emph{Proceedings of the 2021 Conference of the North American
  Chapter of the Association for Computational Linguistics: Human Language
  Technologies}, pages 3002--3017, Online. Association for Computational
  Linguistics.

\bibitem[{Chen and Yang(2020)}]{chen-yang-2020-multi}
Jiaao Chen and Diyi Yang. 2020.
\newblock \href {https://doi.org/10.18653/v1/2020.emnlp-main.336} {Multi-view
  sequence-to-sequence models with conversational structure for abstractive
  dialogue summarization}.
\newblock In \emph{Proceedings of the 2020 Conference on Empirical Methods in
  Natural Language Processing (EMNLP)}, pages 4106--4118, Online. Association
  for Computational Linguistics.

\bibitem[{Chen and Guestrin(2016)}]{chen2016xgboost}
Tianqi Chen and Carlos Guestrin. 2016.
\newblock Xgboost: A scalable tree boosting system.
\newblock In \emph{Proceedings of the 22nd {ACM} sigkdd international
  conference on knowledge discovery and data mining}, pages 785--794.

\bibitem[{Cheng et~al.(2019)Cheng, Fang, and
  Ostendorf}]{cheng-etal-2019-dynamic}
Hao Cheng, Hao Fang, and Mari Ostendorf. 2019.
\newblock \href {https://doi.org/10.18653/v1/N19-1284} {A dynamic speaker model
  for conversational interactions}.
\newblock In \emph{Proceedings of the 2019 Conference of the North {A}merican
  Chapter of the Association for Computational Linguistics: Human Language
  Technologies, Volume 1 (Long and Short Papers)}, pages 2772--2785,
  Minneapolis, Minnesota. Association for Computational Linguistics.

\bibitem[{Core and Allen(1997)}]{Core97codingdialogs}
Mark~G Core and James~F Allen. 1997.
\newblock Coding dialogs with the damsl annotation scheme.
\newblock In \emph{in Proc. Working Notes AAAI Fall Symp. Commun. Action in
  Humans}.

\bibitem[{Devlin et~al.(2019)Devlin, Chang, Lee, and
  Toutanova}]{devlin-etal-2019-bert}
Jacob Devlin, Ming-Wei Chang, Kenton Lee, and Kristina Toutanova. 2019.
\newblock \href {https://doi.org/10.18653/v1/N19-1423} {{BERT}: Pre-training of
  deep bidirectional transformers for language understanding}.
\newblock In \emph{Proceedings of the 2019 Conference of the North {A}merican
  Chapter of the Association for Computational Linguistics: Human Language
  Technologies, Volume 1 (Long and Short Papers)}, pages 4171--4186,
  Minneapolis, Minnesota. Association for Computational Linguistics.

\bibitem[{Efron and Tibshirani(1993)}]{Efron1993AnIT}
Bradley Efron and Robert Tibshirani. 1993.
\newblock An introduction to the bootstrap.
\newblock In \emph{Chapman \& Hall/CRC Monographs on Statistics \& Applied
  Probability.} Taylor \& Francis.

\bibitem[{Grosz and Sidner(1986)}]{grosz-sidner-1986-attention}
Barbara~J. Grosz and Candace~L. Sidner. 1986.
\newblock \href {https://aclanthology.org/J86-3001} {Attention, intentions, and
  the structure of discourse}.
\newblock \emph{Computational Linguistics}, 12(3):175--204.

\bibitem[{Gunasekara et~al.(2019)Gunasekara, Nahamoo, Polymenakos, Ciaurri,
  Ganhotra, and Fadnis}]{GUNASEKARA2019}
R.~Chulaka Gunasekara, David Nahamoo, Lazaros~C. Polymenakos, David~Echeverría
  Ciaurri, Jatin Ganhotra, and Kshitij~P. Fadnis. 2019.
\newblock Quantized dialog – a general approach for conversational systems.
\newblock \emph{Computer Speech and Language}, 54:17--30.

\bibitem[{Gururangan et~al.(2020)Gururangan, Marasovi{\'c}, Swayamdipta, Lo,
  Beltagy, Downey, and Smith}]{gururangan-etal-2020-dont}
Suchin Gururangan, Ana Marasovi{\'c}, Swabha Swayamdipta, Kyle Lo, Iz~Beltagy,
  Doug Downey, and Noah~A. Smith. 2020.
\newblock \href {https://doi.org/10.18653/v1/2020.acl-main.740} {Don{'}t stop
  pretraining: Adapt language models to domains and tasks}.
\newblock In \emph{Proceedings of the 58th Annual Meeting of the Association
  for Computational Linguistics}, pages 8342--8360, Online. Association for
  Computational Linguistics.

\bibitem[{He et~al.(2018)He, Chen, Balakrishnan, and
  Liang}]{he-etal-2018-decoupling}
He~He, Derek Chen, Anusha Balakrishnan, and Percy Liang. 2018.
\newblock \href {https://doi.org/10.18653/v1/D18-1256} {Decoupling strategy and
  generation in negotiation dialogues}.
\newblock In \emph{Proceedings of the 2018 Conference on Empirical Methods in
  Natural Language Processing}, pages 2333--2343, Brussels, Belgium.
  Association for Computational Linguistics.

\bibitem[{Johnson et~al.(2019)Johnson, Douze, and
  J{\'e}gou}]{johnson2019billion}
Jeff Johnson, Matthijs Douze, and Herv{\'e} J{\'e}gou. 2019.
\newblock Billion-scale similarity search with {GPUs}.
\newblock \emph{IEEE Transactions on Big Data}, 7(3):535--547.

\bibitem[{Joshi et~al.(2021)Joshi, Balachandran, Vashishth, Black, and
  Tsvetkov}]{joshi2021dialograph}
Rishabh Joshi, Vidhisha Balachandran, Shikhar Vashishth, Alan Black, and Yulia
  Tsvetkov. 2021.
\newblock \href {https://openreview.net/forum?id=kDnal_bbb-E} {Dialograph:
  Incorporating interpretable strategy-graph networks into negotiation
  dialogues}.
\newblock In \emph{International Conference on Learning Representations}.

\bibitem[{Joty et~al.(2011)Joty, Carenini, and Lin}]{Joty+11}
Shafiq Joty, Giuseppe Carenini, and Chin-Yew Lin. 2011.
\newblock Unsupervised modeling of dialog acts in asynchronous conversations.
\newblock In \emph{Proc.\ International Joint Conference on Artificial
  Intelligence}, pages 1807--1813.

\bibitem[{Jurafsky et~al.(1997)Jurafsky, Shriberg, , and
  Biasca}]{Jurafsky1997SWDA}
Dan Jurafsky, Elizabeth Shriberg, , and Debra Biasca. 1997.
\newblock Switchboard swbd-damsl shallow-discourse-function annotation coders
  manual, draft 13.
\newblock Technical report, University of Colorado, Boulder.

\bibitem[{Lundberg and Lee(2017)}]{NIPS2017_7062}
Scott~M Lundberg and Su-In Lee. 2017.
\newblock \href
  {http://papers.nips.cc/paper/7062-a-unified-approach-to-interpreting-model-predictions.pdf}
  {A unified approach to interpreting model predictions}.
\newblock In I.~Guyon, U.~V. Luxburg, S.~Bengio, H.~Wallach, R.~Fergus,
  S.~Vishwanathan, and R.~Garnett, editors, \emph{Advances in Neural
  Information Processing Systems 30}, pages 4765--4774. Curran Associates, Inc.

\bibitem[{Merialdo(1994)}]{merialdo-1994-tagging}
Bernard Merialdo. 1994.
\newblock \href {https://aclanthology.org/J94-2001} {Tagging {E}nglish text
  with a probabilistic model}.
\newblock \emph{Computational Linguistics}, 20(2):155--171.

\bibitem[{Murphy(2012)}]{Murphy:2012}
Kevin Murphy. 2012.
\newblock \emph{Machine Learning: A Probabilistic Perspective}.
\newblock MIT Press.

\bibitem[{Ostendorf and Singer(1997)}]{ostendorf1997hmm}
Mari Ostendorf and Harald Singer. 1997.
\newblock Hmm topology design using maximum likelihood successive state
  splitting.
\newblock \emph{Computer Speech \& Language}, 11(1):17--41.

\bibitem[{Pappagari et~al.(2019)Pappagari, Zelasko, Villalba, Carmiel, and
  Dehak}]{pappagari2019hierarchical}
Raghavendra Pappagari, Piotr Zelasko, Jes{\'u}s Villalba, Yishay Carmiel, and
  Najim Dehak. 2019.
\newblock Hierarchical transformers for long document classification.
\newblock In \emph{2019 IEEE Automatic Speech Recognition and Understanding
  Workshop (ASRU)}, pages 838--844. IEEE.

\bibitem[{Paszke et~al.(2019)Paszke, Gross, Massa, Lerer, Bradbury, Chanan,
  Killeen, Lin, Gimelshein, Antiga, Desmaison, Kopf, Yang, DeVito, Raison,
  Tejani, Chilamkurthy, Steiner, Fang, Bai, and Chintala}]{NEURIPS2019_9015}
Adam Paszke, Sam Gross, Francisco Massa, Adam Lerer, James Bradbury, Gregory
  Chanan, Trevor Killeen, Zeming Lin, Natalia Gimelshein, Luca Antiga, Alban
  Desmaison, Andreas Kopf, Edward Yang, Zachary DeVito, Martin Raison, Alykhan
  Tejani, Sasank Chilamkurthy, Benoit Steiner, Lu~Fang, Junjie Bai, and Soumith
  Chintala. 2019.
\newblock \href
  {http://papers.neurips.cc/paper/9015-pytorch-an-imperative-style-high-performance-deep-learning-library.pdf}
  {Pytorch: An imperative style, high-performance deep learning library}.
\newblock In H.~Wallach, H.~Larochelle, A.~Beygelzimer, F.~d\textquotesingle
  Alch\'{e}-Buc, E.~Fox, and R.~Garnett, editors, \emph{Advances in Neural
  Information Processing Systems 32}, pages 8024--8035. Curran Associates, Inc.

\bibitem[{Paul(2012)}]{paul-2012-mixed}
Michael~J. Paul. 2012.
\newblock \href {https://aclanthology.org/D12-1009} {Mixed membership {M}arkov
  models for unsupervised conversation modeling}.
\newblock In \emph{Proceedings of the 2012 Joint Conference on Empirical
  Methods in Natural Language Processing and Computational Natural Language
  Learning}, pages 94--104, Jeju Island, Korea. Association for Computational
  Linguistics.

\bibitem[{Ramanath et~al.(2014)Ramanath, Liu, Sadeh, and
  Smith}]{ramanath-etal-2014-unsupervised}
Rohan Ramanath, Fei Liu, Norman Sadeh, and Noah~A. Smith. 2014.
\newblock \href {https://doi.org/10.3115/v1/P14-2099} {Unsupervised alignment
  of privacy policies using hidden {M}arkov models}.
\newblock In \emph{Proceedings of the 52nd Annual Meeting of the Association
  for Computational Linguistics (Volume 2: Short Papers)}, pages 605--610,
  Baltimore, Maryland. Association for Computational Linguistics.

\bibitem[{Rasley et~al.(2020)Rasley, Rajbhandari, Ruwase, and
  He}]{rasley2020deepspeed}
Jeff Rasley, Samyam Rajbhandari, Olatunji Ruwase, and Yuxiong He. 2020.
\newblock Deepspeed: System optimizations enable training deep learning models
  with over 100 billion parameters.
\newblock In \emph{Proceedings of the 26th ACM SIGKDD International Conference
  on Knowledge Discovery \& Data Mining}, pages 3505--3506.

\bibitem[{Ritter et~al.(2010)Ritter, Cherry, and
  Dolan}]{ritter-etal-2010-unsupervised}
Alan Ritter, Colin Cherry, and Bill Dolan. 2010.
\newblock \href {https://aclanthology.org/N10-1020} {Unsupervised modeling of
  {T}witter conversations}.
\newblock In \emph{Human Language Technologies: The 2010 Annual Conference of
  the North {A}merican Chapter of the Association for Computational
  Linguistics}, pages 172--180, Los Angeles, California. Association for
  Computational Linguistics.

\bibitem[{Santra et~al.(2021)Santra, Anusha, and
  Goyal}]{santra-etal-2021-hierarchical}
Bishal Santra, Potnuru Anusha, and Pawan Goyal. 2021.
\newblock \href {https://doi.org/10.18653/v1/2021.naacl-main.449} {Hierarchical
  transformer for task oriented dialog systems}.
\newblock In \emph{Proceedings of the 2021 Conference of the North American
  Chapter of the Association for Computational Linguistics: Human Language
  Technologies}, pages 5649--5658, Online. Association for Computational
  Linguistics.

\bibitem[{Schreiber(2018)}]{schreiber2018pomegranate}
Jacob Schreiber. 2018.
\newblock Pomegranate: fast and flexible probabilistic modeling in python.
\newblock \emph{Journal of Machine Learning Research}, 18(164):1--6.

\bibitem[{Shang et~al.(2018)Shang, Ding, Zhang, Tixier, Meladianos,
  Vazirgiannis, and Lorr{\'e}}]{shang-etal-2018-unsupervised}
Guokan Shang, Wensi Ding, Zekun Zhang, Antoine Tixier, Polykarpos Meladianos,
  Michalis Vazirgiannis, and Jean-Pierre Lorr{\'e}. 2018.
\newblock \href {https://doi.org/10.18653/v1/P18-1062} {Unsupervised
  abstractive meeting summarization with multi-sentence compression and
  budgeted submodular maximization}.
\newblock In \emph{Proceedings of the 56th Annual Meeting of the Association
  for Computational Linguistics (Volume 1: Long Papers)}, pages 664--674,
  Melbourne, Australia. Association for Computational Linguistics.

\bibitem[{Shi et~al.(2019)Shi, Zhao, and Yu}]{shi-etal-2019-unsupervised}
Weiyan Shi, Tiancheng Zhao, and Zhou Yu. 2019.
\newblock \href {https://doi.org/10.18653/v1/N19-1178} {Unsupervised dialog
  structure learning}.
\newblock In \emph{Proceedings of the 2019 Conference of the North {A}merican
  Chapter of the Association for Computational Linguistics: Human Language
  Technologies, Volume 1 (Long and Short Papers)}, pages 1797--1807,
  Minneapolis, Minnesota. Association for Computational Linguistics.

\bibitem[{Veli{\v{c}}kovi{\'c} et~al.(2018)Veli{\v{c}}kovi{\'c}, Cucurull,
  Casanova, Romero, Lio, and Bengio}]{velivckovic2017graph}
Petar Veli{\v{c}}kovi{\'c}, Guillem Cucurull, Arantxa Casanova, Adriana Romero,
  Pietro Lio, and Yoshua Bengio. 2018.
\newblock \href {https://openreview.net/forum?id=rJXMpikCZ} {Graph attention
  networks}.
\newblock In \emph{International Conference on Learning Representations}.

\bibitem[{Wolf et~al.(2020)Wolf, Debut, Sanh, Chaumond, Delangue, Moi, Cistac,
  Rault, Louf, Funtowicz, Davison, Shleifer, von Platen, Ma, Jernite, Plu, Xu,
  Le~Scao, Gugger, Drame, Lhoest, and Rush}]{wolf-etal-2020-transformers}
Thomas Wolf, Lysandre Debut, Victor Sanh, Julien Chaumond, Clement Delangue,
  Anthony Moi, Pierric Cistac, Tim Rault, Remi Louf, Morgan Funtowicz, Joe
  Davison, Sam Shleifer, Patrick von Platen, Clara Ma, Yacine Jernite, Julien
  Plu, Canwen Xu, Teven Le~Scao, Sylvain Gugger, Mariama Drame, Quentin Lhoest,
  and Alexander Rush. 2020.
\newblock \href {https://doi.org/10.18653/v1/2020.emnlp-demos.6} {Transformers:
  State-of-the-art natural language processing}.
\newblock In \emph{Proceedings of the 2020 Conference on Empirical Methods in
  Natural Language Processing: System Demonstrations}, pages 38--45, Online.
  Association for Computational Linguistics.

\bibitem[{Zhai and Williams(2014)}]{zhai-williams-2014-discovering}
Ke~Zhai and Jason~D. Williams. 2014.
\newblock \href {https://doi.org/10.3115/v1/P14-1004} {Discovering latent
  structure in task-oriented dialogues}.
\newblock In \emph{Proceedings of the 52nd Annual Meeting of the Association
  for Computational Linguistics (Volume 1: Long Papers)}, pages 36--46,
  Baltimore, Maryland. Association for Computational Linguistics.

\bibitem[{Zhou et~al.(2020)Zhou, Tsvetkov, Black, and Yu}]{Zhou2020Augmenting}
Yiheng Zhou, Yulia Tsvetkov, Alan~W Black, and Zhou Yu. 2020.
\newblock \href {https://openreview.net/forum?id=ryxQuANKPB} {Augmenting
  non-collaborative dialog systems with explicit semantic and strategic dialog
  history}.
\newblock In \emph{International Conference on Learning Representations}.

\end{thebibliography}
